\RequirePackage{fix-cm}
\documentclass[twocolumn]{svjour3}          %
\smartqed  %
\usepackage{graphicx}
\usepackage{mathptmx}      %

\usepackage{amsmath}
\usepackage{amssymb} %
\usepackage{color}

\usepackage{times} 
\usepackage{url} 
\usepackage{epsfig}   
\usepackage{subfig}   
\usepackage{mwe}

\usepackage{algorithm} 
\usepackage{breakcites}
\usepackage{microtype} %
\usepackage[noend]{algpseudocode} 
\usepackage{hhline} 
\usepackage[sort]{natbib} %
\renewcommand{\paragraph}[1]{\smallskip\noindent{\bf{#1}}}
\usepackage{afterpage}

\DeclareRobustCommand{\eg}{e.g.\@\xspace}
\DeclareRobustCommand{\ie}{i.e.\@\xspace}
\DeclareRobustCommand{\etc}{etc.\@\xspace}

\begin{document}

\title{Compact Deep Aggregation for Set Retrieval}

\titlerunning{Compact Deep Aggregation for Set Retrieval}        %

\author{\author{Yujie Zhong \and
Relja Arandjelovi\'c \and
Andrew Zisserman}
}

\authorrunning{Y. Zhong, R. Arandjelovi\'c and A. Zisserman} %

\institute{
Y. Zhong and A. Zisserman \at
Visual Geometry Group, Department of Engineering Science, University of Oxford, UK \\
\email{\{yujie,az\}@robots.ox.ac.uk} \and
R. Arandjelovi\'c \at
DeepMind\\
\email{relja@google.com}}

\maketitle

\begin{abstract}

The objective of this work is to learn a compact embedding of a set of
descriptors that is suitable for efficient retrieval and ranking,
whilst maintaining discriminability of the individual descriptors.  We
focus on a specific example of this general problem -- that of
retrieving images containing multiple faces from a large scale dataset
of images. Here the set consists of the face descriptors in each
image, and given a query for multiple identities, the goal is then to
retrieve, in order, images which contain all the identities, all but
one, \etc

To this end, we make the following contributions: first, we propose a
CNN architecture -- {\em SetNet} -- to achieve the objective: it
learns face descriptors and their aggregation over a set to produce a
compact fixed length descriptor designed for set retrieval, and the
score of an image is a count of the number of identities that match
the query; second, we show that this compact descriptor has minimal
loss of discriminability up to two  faces per image, and degrades
slowly after that -- far exceeding a number of baselines; third, we
explore the speed vs.\ retrieval quality trade-off 
for set retrieval using this compact descriptor; and, finally, we collect and
annotate a large dataset of images containing various number of
celebrities, which we use for evaluation and is publicly
released.

\end{abstract}

\section{Introduction}\label{sec:intro}
Suppose we wish to retrieve all images in a
very large collection of personal photos that contain a particular set of people, such
as a group of friends or a family. Then we would like the retrieved
images that contain all of the set to be ranked first, followed by
images containing subsets, \eg if there are three friends in the query, 
then first would be images containing all three friends, then images
containing two of the three, followed by images containing only one of them.
Consider another example where a fashion savvy customer is researching
a dress and a handbag.
A useful retrieval system would first retrieve images 
in which models or celebrities wear both items, so that the customer can
see if they match.
Following these, it would also be beneficial to show images containing any one of the 
query items so that the customer can discover other fashionable combinations
relevant to their interest.
For these systems to be of practical use,
we would also like this retrieval to happen in real time.

\begin{figure*}[t]
   \begin{center}
         \includegraphics[width=0.95\textwidth]{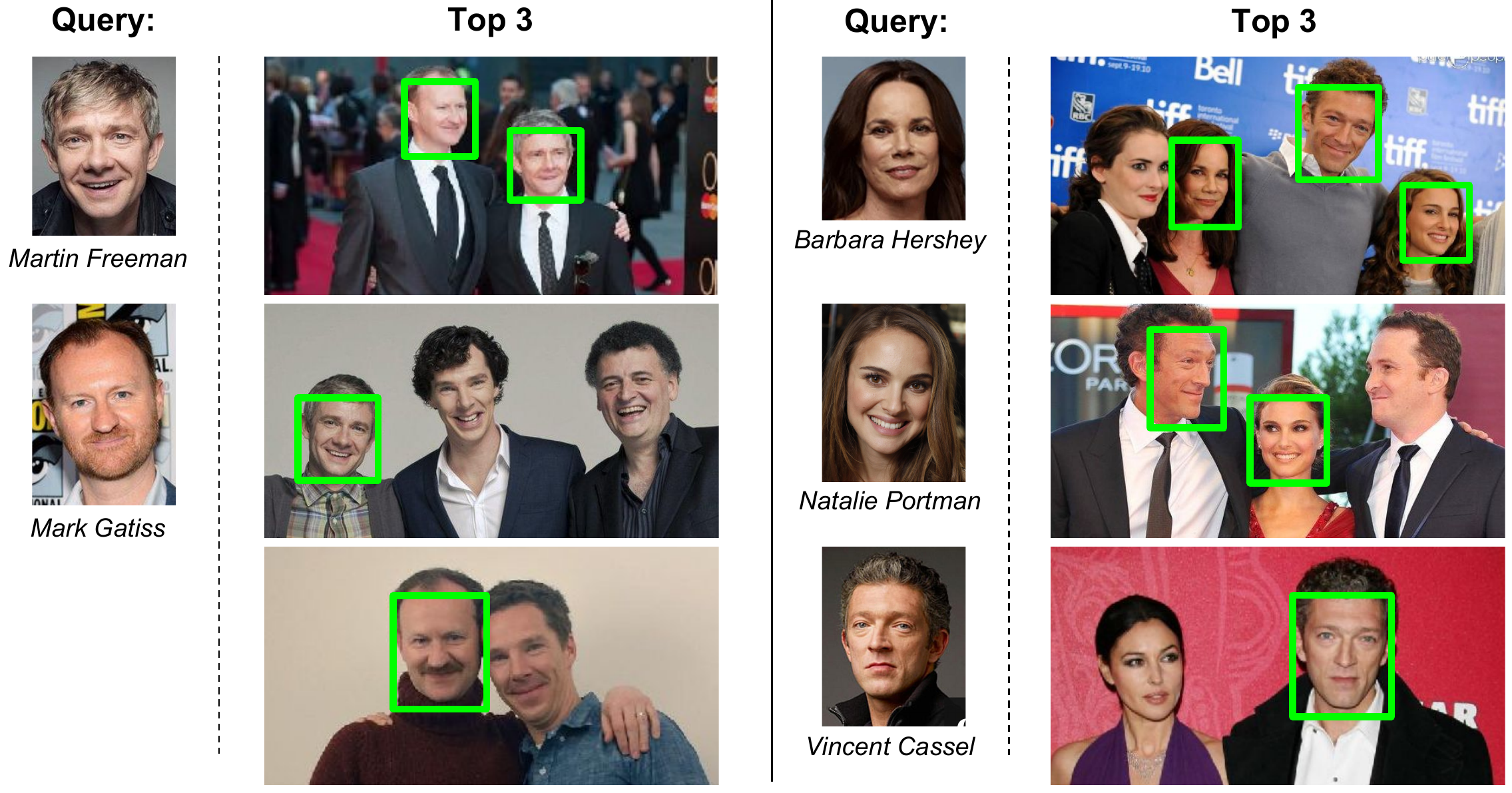}
   \end{center}
   \caption{ \textbf{Images ranked using set retrieval for two example queries.} 
The query faces are given on the left of each example column, 
together with their names (only for reference). 
Left: a query for two identities; right: 
a query for three  identities. The first ranked image in each case contains all the
faces in the query. Lower ranked images partially satisfy the query, and  contain
progressively fewer faces of the query. The results are obtained using the 
compact set retrieval descriptor generated by the SetNet architecture, by searching over 200k images
of the {\em Celebrity Together} dataset introduced  in this paper.}
\label{fig:teaser}
 \end{figure*}

These are examples of a {\em set retrieval problem}: 
the database consists of many sets of elements
(\eg the set is an image containing  multiple faces, or of a person with multiple fashion items),
and we wish to order the sets according to a query
(\eg multiple identities, or multiple fashion items)
such that those sets satisfying the query completely 
are ranked first 
(\ie those images that contain all the identities of the query, or all the query fashion items), 
followed by sets that satisfy all but one of
the query elements, \etc  
An example of this ranking for the face retrieval problem is 
shown in Fig.~\ref{fig:teaser} for two queries.

In this work, we focus on the problem of 
retrieving a set of identities,
but the same approach can be used for other set retrieval problems
such as the aforementioned fashion retrieval.
We can operationalize this by scoring each face in each photo of the collection
as to whether they are one of the identities in the query.
Each face is represented by a fixed length vector,
and identities are scored by logistic regression classifiers.
But, consider the situation where the dataset is very large, containing
millions or billions of images each containing multiple faces. In this situation two aspects are
crucial for real time retrieval: first, that all operations take place in memory (not reading from disk), and second
that an efficient algorithm is used when searching 
for images that satisfy the query. The problem is that storing a fixed length vector for each {\em face} in memory
is prohibitively expensive at this scale, but this cost can be significantly reduced if a fixed length vector 
is only stored for each {\em set of faces} in an image (since there are far fewer images than faces). As well as
reducing the memory cost this also reduces the run time cost of the search since fewer vectors need to be
scored.

So, the question we investigate in this paper is the following: can we aggregate the
{\em set of vectors} representing the multiple faces in an image into
a {\em single vector} with little loss of set-retrieval performance? If so, then
the cost of both memory and retrieval can be significantly reduced as
only one vector per {\em image} (rather than one per {\em face}) have
to be stored and scored.  

Although we have motivated this question by face retrieval it is quite
general: there is a set of elements (which are permutation-invariant), 
each element is represented by a vector of
dimension $D_e$, and we wish to represent this set by a single vector of
dimension $D$, where $D = D_e$ in practice, without losing information
essential for the task.  Of course, if the total number of elements in all sets,
$N$, is such that $N \leq D$ then this certainly can be achieved
provided that the set of vectors are orthogonal. However, we will
consider the situation commonly found in practice
where $N \gg D$, \eg $D$ is small, typically 128 (to keep the memory
footprint low), and  $N$ is in the thousands.

We make the following contributions: first, we introduce a trainable CNN  architecture 
for the set-retrieval task that is able to learn to aggregate face vectors into a fixed length descriptor
in order to minimize interference, and
also is able to rank the face sets according to how many identities are in common with the query
using this descriptor.
To do this, 
we propose a paradigm shift where we draw motivation
from image retrieval based on local descriptors.
In image retrieval, it is common practice to aggregate all local descriptors
of an image into a fixed-size image-level vector representation,
such as bag-of-words~\citep{Sivic03} and VLAD~\citep{Jegou10};
this brings both memory and speed improvements over storing all local
descriptors individually.
We generalize this concept to set retrieval, where instead of aggregating
local interest point descriptors, set element descriptors
are pooled into a single fixed-size set-level representation.
For the particular case of face set retrieval, this corresponds to aggregating
face descriptors into a set representation.
The novel aggregation procedure is described in Sec.~\ref{sec:compact}
where the aggregator can be trained in an end-to-end manner with 
different feature extractor CNNs.

Our second contribution is to 
introduce  a dataset  annotated with
multiple faces per images. In Sec.~\ref{sec:dataset} we describe a
pipeline for automatically generating a labelled dataset of pairs (or
more) of celebrities per image. This {\em Celebrity Together}  dataset
contains around 200k images with more than half a million faces in
total. It is publicly released.

The performance of the set-level descriptors is evaluated in Sec.~\ref{sec:exp}. We first
`stress test' the descriptors by progressively increasing the number of faces in each set, and
monitoring their retrieval performance.  We also evaluate retrieval on the 
 {\em Celebrity Together}  dataset, where images contain a variable number of faces, with many
not corresponding to the queries, and explore efficient algorithms that can achieve immediate
(real-time) retrieval on very large scale datasets.
Finally, we investigate what the trained network learns by analyzing
why the face descriptors learnt by the network are  well suited for aggregation.

Note, although we have grounded the set retrieval problem as faces,
the treatment is quite general: it only assumes that dataset elements
are represented by vectors and the scoring function is a scalar
product. We return to this point in the conclusion.

\section{Related work}
\label{sec:review}

To the best of our knowledge, this paper is the first to consider
the set retrieval problem.
However, the general area of image
retrieval has an extensive literature that we build on here.

\subsection{Aggregating descriptors}
One of the central problems that has been studied in large scale image
instance 
retrieval is how to condense the information stored in
multiple local descriptors such as SIFT~\citep{Lowe04},
into a single compact vector to represent the image.
This problem has been driven by the need
to keep the memory footprint low for very large image datasets.
An early approach is to cluster descriptors into visual words
and represent the image as a histogram of word occurrences --
bag-of-visual-words~\citep{Sivic03}.
Performance can be improved by aggregating
local descriptors within each cluster, in representations such as
Fisher Vectors~\citep{Perronnin10,Jegou11b} 
and VLAD~\citep{Jegou10}.
In particular, VLAD --  `Vector of Locally
Aggregated Descriptors' by~\cite{Jegou10}
and its improvements~\citep{Delhumeau13,Arandjelovic13,Jegou14}
was used to obtain very compact descriptions via
dimensionality reduction by PCA,
considerably reducing the memory
requirements compared to earlier bag-of-words based 
methods~\citep{Sivic03,Nister06,Philbin07,Chum07b}.
VLAD has 
superior ability in maintaining the information about individual local
descriptors while performing aggressive dimensionality reduction.

VLAD has recently been adapted into a
differentiable CNN layer, NetVLAD~\citep{Arandjelovic16},
making it end-to-end trainable.
We incorporate a modified form of
the NetVLAD layer in our SetNet architecture. An alternative, but
related, very recent approach is the memory vector formulation
proposed by \cite{Iscen17}, but we have not employed it here as it has not been
made differentiable yet.

\subsection{Image retrieval}
Another strand of research we build on is category level retrieval,
where in our case the category is a face.
This is another classical area of interest with many related
works~\citep{Wang10,Zhou10,Torresani10,Perronnin10a,Chatfield11}.
Since the advent of deep
CNNs~\citep{Krizhevsky12}, the standard solution
has been to use features from the last hidden layer of a CNN
trained on ImageNet~\citep{Donahue13,Sermanet13,Razavian14}
as the image representation.
This has enabled  a very successful approach for retrieving images
containing the query category by training linear SVMs on-the-fly
and ranking images based on their classification
scores\citep{Chatfield14a,Chatfield15}.
For the case of faces, the
feature vector is produced from the face region using a CNN
trained to classify or embed faces~\citep{Parkhi15,Schroff15,Taigman14}.
\cite{Li11} tackle face retrieval 
by using binary codes that jointly encode
identity discriminability and a number 
of facial attributes. 
\cite{Li15} extend image-based face retrieval 
to video scenarios, and learn low-dimensional 
binary vectors encoding the face tracks 
and hierarchical video representations.

Apart from single-object retrieval, there are also 
works on compound query retrieval, such as~\citep{Zhong16}
who focus on finding a specific face in a place.
While this can be seen as set retrieval where each set has exactly two
elements, one face and one place, it is not applicable to our problem
as their elements come from different sources (two different networks)
and are never combined together in a set-level representation.

\subsection{Approaches for sets}
Also relevant are works that explicitly deal with sets of vectors.
\cite{Kondor03} developed a kernel between
vector sets by characterising each set as a Gaussian in some Hilbert space.
However, their set representation cannot currently be combined with CNNs and
trained in an end-to-end fashion.
\cite{Feng17} 
learn a compact set 
representation for matching of image sets by jointly optimizing a
neural network and hashing functions 
in an end-to-end fashion.
Pooling architectures are commonly used to deal with sets,
\eg for combining information from a set of views  for 3D shape recognition~\citep{Shi15,Su15}.
\cite{Zaheer17} investigate
permutation-invariant objective functions for operating on sets,
although their method boils down to average pooling of input vectors,
which we compare to as a baseline.
Some works adopt attention mechanism 
~\citep{Vinyals16,Yang18,Ilse18,Lee18}
or relational network~\citep{Santoro17}
into set operations to model the 
interactions between elements in the input set.
\cite{Rezatofighi17} consider the problem of predicting
sets, \ie having a network which outputs sets, rather than our case where
a set of elements is an input to be processed and described
with a single vector.
\cite{Siddiquie11} formulate a face 
retrieval task where the query is a set of attributes.
\cite{Yang17} and \cite{Liu17} learn 
to aggregate set of faces of the same identity
for recognition, 
with different weightings based on image quality 
using an attention mechanism.
For the same task, \cite{Wang17a} represent face image 
sets as covariance matrices.
Note, our objective differs from these: we are not aggregating sets of 
faces of the same identity, but instead aggregating sets of faces of different identities.

\section{SetNet -- a CNN for set retrieval}
\label{sec:compact}

As described in the previous section, using a single fixed-size vector
to represent a set of vectors is a highly appealing approach due to
its superior speed and memory footprint over storing a descriptor-per-element.
In this section, we  propose  a CNN architecture,  \emph{SetNet}, 
for the end-task of set retrieval. There are two objectives:

\paragraph{1.}
It should learn the element descriptors together with the aggregation in order to minimise the loss in
face classification performance between using individual descriptors for each face, and an aggregated
descriptor for the set of faces. This is achieved by training the network for this task, using an architecture 
combining ResNet for the individual descriptors together with NetVLAD for the aggregation.

\paragraph{2.} 
It should be able to rank the images using the 
aggregated descriptor in order of the number of faces in
each image that correspond to the identities in the query. 
This is achieved by scoring each face using a logistic regression classifier.
Since the score of each classifier lies between 0 and 1, the score for the image can simply be computed as the sum of the individual scores, and this summed score determines the ranking function.

As an example of the scoring function, if the search is for two
identities and an image contains faces of both of them (and maybe
other faces as well), then the ideal score for each relevant face would be one,
and the sum of scores for the image would be two. If an image only
contains one of the identities, then the sum of the scores would be
one. The images with higher summed scores are then ranked higher and
this naturally orders the images by the number of faces it contains
that satisfy the query.

\paragraph{Network Deployment.}
To deploy  the set level descriptor for retrieval in a large scale dataset, there are two stages:

\noindent {\bf Offline:} SetNet is used to compute
face descriptors for each face in an image, and aggregate them to generate a set-vector representing
the image. This procedure is carried out for every image in the dataset, so that each image is represented
by a single vector.

\noindent At {\bf run-time}, to search for an identity,  a face 
descriptor is  computed for the query face using SetNet, and 
a logistic regression classifier used to score each image based on the scalar product between its set-vector
and the query face descriptor.
Searching with a set of identities amounts to summing up the image scores of
each query identity.

\begin{figure*}[!t]
   \begin{center}
         \includegraphics[width=0.90\textwidth]{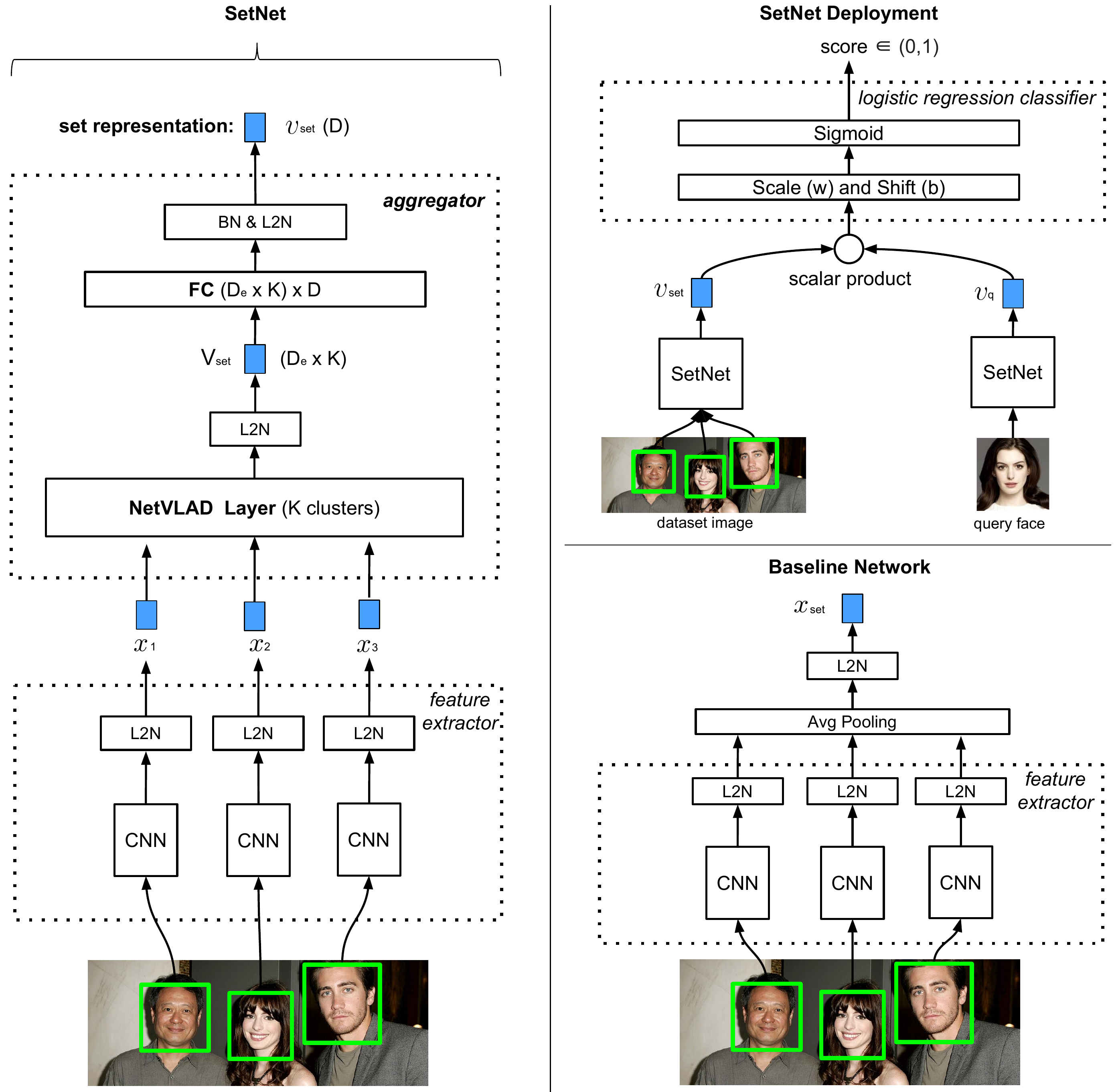}
   \end{center}
   \caption{ \textbf{SetNet architecture and training.}
\textbf{Left}: SetNet -- features are extracted 
from each face in an image using a modified
ResNet-50 or SENet-50. They are aggregated 
using a modified NetVLAD layer into
a single $D_e \times K$ dimensional vector which is 
then reduced to $D$ dimension (where $D$ is 128 or 256)
via a fully connected dimensionality reduction layer,
and L2-normalized to obtain the final image-level
compact representation. 
\textbf{Right (top)}: at test time, a query descriptor, $v_q$,  is obtained
for each query face using SetNet (the face is 
considered as a single-element set),
and the dataset image is scored by a logistic regression classifier on
the scalar product between the 
query descriptor $v_q$ and image set descriptor $v_{set}$.
The final score of an image is then obtained 
by summing the scores of all the query identities.
\textbf{Right (bottom)}: the baseline network 
has the same feature extractor as SetNet, 
but the feature aggregator uses 
an average-pooling layer, rather than NetVLAD.
}
    \label{fig:cnn}
 \end{figure*}

\subsection{SetNet architecture}\label{sec:arc_detail}

In this section we introduce our CNN architecture, designed to aggregate
multiple element (face) descriptors into a single fixed-size set representation.
The \emph{SetNet} architecture (Fig.~\ref{fig:cnn}) conceptually has two parts:
(i) each face is passed through a feature extractor network
separately, producing one descriptor per face;
(ii) the multiple face descriptors are aggregated into a single compact vector
using a modified NetVLAD layer,  followed by a trained dimensionality reduction.
At training time, we add a third part which emulates the run-time
use of logistic regression classifiers.
All three parts of the architecture are described in more detail next.

\paragraph{Feature extraction.}
The first part is a CNN for feature extraction which produces
individual element descriptors ($x_1$ to $x_F$),
where $F$ is the number of faces in an image.
We experiment with two different base architectures --
modified ResNet-50~\citep{He16} and modified SENet-50~\citep{Hu18}, both
chopped after the global average pooling layer.
The ResNet-50 and SENet-50
are modified to produce $D_e$ dimensional 
vectors (where $D_e$ is 128 or 256) in order to keep
the dimensionality of our feature vectors relatively low, and
we have not observed a significant drop in face recognition
performance from the original 2048-D descriptors.
The modification
is implemented by adding a fully-connected (FC) 
layer of size $2048 \times$ $D_e$ after the global 
average pooling layer of the original network, 
in order to obtain a lower $D_e$-dimensional face descriptor.
This additional fully-connected layer essentially acts as 
a dimensionality reduction layer.
This modification introduces around 260k 
additional weights to the network for $D_e = 128$,
and 520k weights for $D_e = 256$,
which is negligible compared to the total number of parameters
of the base networks (25M for ResNet-50, 28M for SENet-50).

\begin{figure}[t]
   \begin{center}
         \includegraphics[width=0.99\columnwidth]{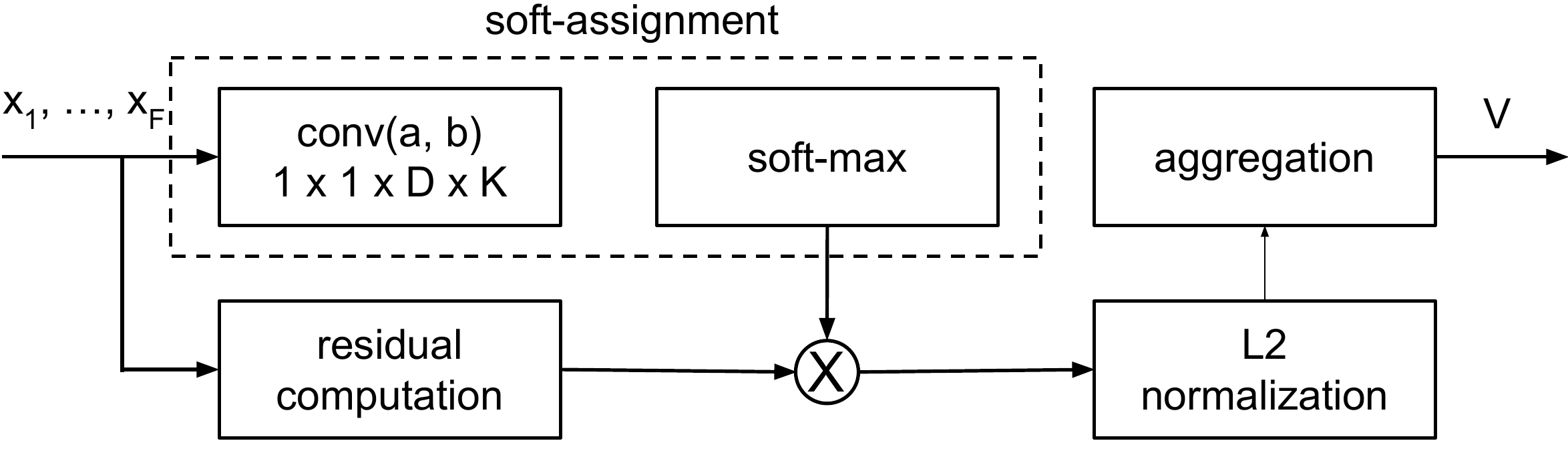}
   \end{center}
   \caption{\textbf{NetVLAD layer.}
Illustration of the NetVLAD layer~\citep{Arandjelovic16}, corresponding
to equation \eqref{eq:netvlad}, and slightly modified
to perform L2-normalization before aggregation;
see Sec.~\ref{sec:netvlad} for details.}
    \label{fig:NetVLAD}
 \end{figure}

\paragraph{Feature aggregation.}
Face features are aggregated into a single vector $V$ using a 
NetVLAD layer (illustrated in Fig.~\ref{fig:NetVLAD}, and described below in Sec.~\ref{sec:netvlad}).
The NetVLAD layer is slightly modified by adding an additional
L2-normalization step -- the total contribution of each face descriptor
to the aggregated sum (\ie its weighted residuals) is L2-normalized
in order for each face descriptor to contribute equally to the final vector;
this procedure is an adaptation of residual normalization~\citep{Delhumeau13}
of the vanilla VLAD to NetVLAD.
The NetVLAD-pooled features are reduced 
to be $D$-dimensional by means of a
fully-connected layer followed by batch-normalization
~\citep{Ioffe15},
and L2-normalized to produce 
the final set representation $v_{set}$.

\paragraph{Training block.}
At training time, an additional logistic regression loss layer is added to
mimic  the run-time scenario where 
a logistic regression classifier is used to score each image based on the scalar product between its set-vector
and the query face descriptor. Note, SetNet is used to generate both the set-vector and the face descriptor.
Sec.~\ref{sec:loss} describes the appropriate loss and training procedure
in more detail.

\subsection{NetVLAD trainable pooling}
\label{sec:netvlad}

NetVLAD  has been shown to outperform sum and max pooling for the same
vector dimensionality, which makes
it perfectly suited for our task.
Here we provide a brief overview of NetVLAD, for full details please refer 
to~\citep{Arandjelovic16}. 

For $F$ $D$-dimensional input descriptors $\{x_i\}$ and
a chosen number of clusters $K$, NetVLAD pooling produces a single $D \times K$
vector $V$ (for convenience written as a $D \times K$ matrix) according
to the following equation:
\begin{equation}
V(j,k) = \sum_{i=1}^{F} \frac{e^{a_k^Tx_i + b_k}}{\sum_{k'}{e^{a_{k'}^Tx_i + b_{k'}}}} (x_i(j) - c_k(j)), \quad j\in 1, \dots, D
\label{eq:netvlad}
\end{equation}
where $\{a_k\}$, $\{b_k\}$ and $\{c_k\}$ are trainable parameters for $k \in [1, 2, \dots, K]$.
The first term corresponds to the soft-assignment weight
of the input vector $x_i$ for cluster $k$,
while the second term computes the residual between the vector and
the cluster centre.
Finally, the vector is L2-normalized.

\subsection{Loss function and training procedure}
\label{sec:loss}
In order to achieve the two objectives outlined at the beginning of
Sec.~\ref{sec:compact}, a Multi-label
logistic regression loss is used.
Suppose a particular training image contains $F$ faces, and the mini-batch consists of faces
for $P$ identities. Then in a
forward pass at training time, the descriptors for the $F$ faces are
aggregated into a single feature vector, $v_{set}$, using the SetNet
architecture, and a face descriptor,  $v_{f}$,  is computed using SetNet for each of the faces of the $P$ identities.
The training image is then scored for each face $f$ by applying a logistic regressor classifier to
the scalar product $v_{set}^T v_f $, and the score should ideally be one for each 
of  the $F$ identities in the image, and zero for the other $P - F$ faces.
The loss measures the deviation from this ideal score, and the network learns to achieve
this by maintaining the discriminability for individual face descriptors after aggregation.

In detail, incorporating the loss  is achieved by adding an additional layer at training time
which contains a logistic regression loss for each of the $P$ training identities,
and is trained together with the rest of the network.

\paragraph{Multi-label logistic regression loss.}
For each training image (set of faces), 
the loss $L$ is computed as:

\begin{equation}
\begin{split}
	L = -\sum^P_{f=1}
	y_f \log(\sigma(w(v_f^T v_{set})+b))
	+ \\
	(1-y_f) \log(1-\sigma(w(v_f^T v_{set})+b))
\label{eq:loss}
\end{split}
\end{equation}
where $\sigma(s) = 1/(1+\exp(-s))$ is a logistic function,
$P$ is the number of face descriptors
(the size of the mini-batches at training), and 
$w$ and $b$ are the scaling factor and shifting bias respectively of the logistic regression classifier,
and $y_f$ is a binary indicator whether face $f$ is in the image or not.
Note that multiple $y_f$'s are equal to 1 
if there are multiple faces which correspond to 
the identities in the image.

\subsection{Implementation details}
\label{sec:training}

This section gives full details of the training procedure, including 
how the network is used  at run-time to rank the 
dataset  images given query examples.

\paragraph{Training data.}
The network is trained using faces from the training partition of the 
VGGFace2 dataset~\citep{Cao18}. This consists
of  8631 identities,  with on average  360 face samples for each identity.

\paragraph{Balancing positives and negatives.}
For each training image (face set)  there are many more negatives (the $P- F$ identities outside of the image set)
than positives (identities in the set),
\ie most $y_f$'s in eq.~\eqref{eq:loss} are equal to 0 with only a few 1's.
To restore balance, the contributions of the positives and negatives
to the loss function is down-weighted by their respective counts.

\paragraph{Initialization and pre-training.}
A good (and necessary) initialization for the network is obtained as follows.
The face feature extraction block is pretrained for single
face classification 
on the VGGFace2 Dataset~\citep{Cao18}
using softmax  loss.
The NetVLAD layer, with $K=8$ clusters,
is initialized using k-means as in~\citep{Arandjelovic16}.
The fully-connected layer, 
used to reduce the NetVLAD dimensionality to $D$,
is initialized by PCA, \ie by arranging the first $D$ principal components
into the weight matrix.
Finally, the entire SetNet is trained for face aggregation using the
Multi-label logistic regression loss (Sec.~\ref{sec:loss}).

\paragraph{Training details.}
Training requires face set descriptors computed for each image, and query faces (which may or may not
occur in the image).
The network is trained on 
synthetic face sets which 
are built by randomly sampling identities (e.g.\ two identities per image).
For each identity in a synthetic set,
two faces are randomly sampled (from the average of 360 for each identity): one 
contributes to the set descriptor (by  combining it with samples of the other identities), 
the other is used as a query face, and its scalar product is computed with
all the set descriptors in the same mini-batch.
In our experiments, each mini-batch contains  84  faces.
Stochastic gradient descent
is used to train the network (implemented in MatConvNet~\citep{Vedaldi15},
with weight decay 0.001, momentum 0.9,
and an initial learning rate of 0.001 for pre-training and 0.0001 for fine-tuning;
the learning rates are divided by 10 in later epochs.

Faces are detected from images using MTCNN~\citep{Zhang16}.
Training faces are resized such that the smallest dimension is 256
and random $224 \times 224$ crops are used as inputs to the network.
To further augment the training faces, random horizontal flipping and up to 10 degree rotation
is performed.
At test time, faces are resized so that the smallest dimension is 224
and the central crop is taken.

\paragraph{Dataset retrieval.}
Suppose we wish to retrieve images containing multiple identities  (or a subset of these).
First,  a  face descriptor is produced by \emph{SetNet} 
for each query face.
The face
descriptors are then used to score a dataset  image for each query identity,
followed by summing the individual logistic regression
scores to produce the final image score.
A ranked list is obtained by sorting the dataset  images in
non-increasing score order.
In the case where multiple face examples 
are available for a query identity,
\emph{SetNet} is used to extract face descriptors 
and aggregate 
them to form a richer descriptor for that query identity.

\section{`Celebrity Together' dataset}
\label{sec:dataset}

\begin{figure*}[t]
\vspace{-0mm}
   \begin{center}
          \includegraphics[width=0.99\linewidth]{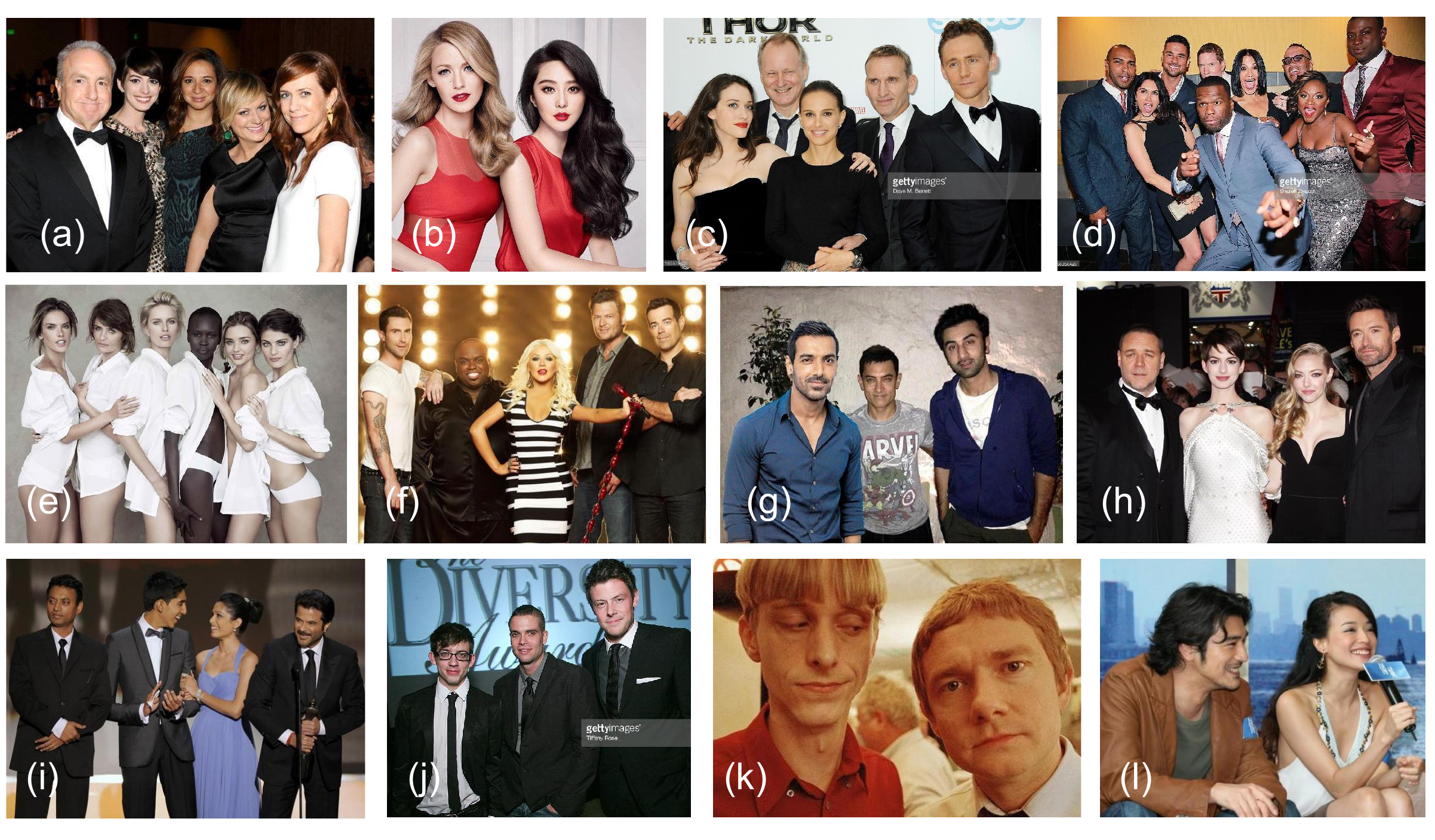}
   \end{center}
   \caption{\textbf{Example images of the {\em Celebrity Together} Dataset.}
Note that only those celebrities who appear in the VGG Face Dataset 
are listed, as the rest are labelled as `unknown'.   
   (a) Amy Poehler, Anne Hathaway, Kristen Wiig, Maya Rudolph.
   (b) Bingbing Fan, Blake Lively. 
   (c) Kat Dennings, Natalie Portman, Tom Hiddleston. 
   (d) Kathrine Narducci, Naturi Naughton, Omari Hardwick, Sinqua Walls. 
	     (e) Helena Christensen, Karolina Kurkova, Miranda Kerr. 
	     (f) Adam Levine, Blake Shelton, CeeLo Green. 
	     (g) Aamir Khan, John Abraham, Ranbir Kapoor.
	     (h) Amanda Seyfried, Anne Hathaway, Hugh Jackman. 
        (i) Anil Kapoor, Irrfan Khan. 
        (j) Cory Monteith, Kevin McHale, Mark Salling. 
        (k) Martin Freeman, Mackenzie Crook. 
	(l) Takeshi Kaneshiro, Qi Shu. 
Additional examples of the dataset images are given in the supplementary material.
}
\vspace{-5mm} 
	\label{fig:example}
 \end{figure*}

A new dataset, {\em Celebrity Together}, is collected and annotated.
It contains images that portray multiple celebrities simultaneously
(Fig.~\ref{fig:example} shows a sample),
making it ideal for testing set retrieval methods.
Unlike the other face datasets, which exclusively contain individual 
face crops, {\em Celebrity Together} is made of full images with multiple
labelled faces.
It contains 194k images and 546k faces in total,
averaging 2.8 faces per image.
The image collection and annotation procedures are explained next.

\subsection{Image collection and annotation procedure}
The dataset is created with the aim of containing multiple people per image,
which makes the image collection procedure much more complex than when
building a single-face-per-image dataset, such as~\citep{Parkhi15}.
The straightforward strategy of~\citep{Parkhi15}, which involves
simply searching for celebrities on an online image search engine,
is inappropriate:
consider an example of searching for Natalie Portman on Google Image Search
-- \emph{all} top ranked images contain only her and no other person.
Here we explain how to overcome this barrier to collect
images with multiple celebrities in them.

\paragraph{Query selection.}
In order to acquire celebrity name pairs, 
we first perform a search by using `\texttt{seed-celebrity} and', 
and then scan the meta information for other celebrity names to produce
a list of (seed-celebrity, celebrity-friend) pairs.
However, one important procedure which is not included in the main paper 
is that some celebrity may have a strong connection to one or more 
particular celebrities, which could prevent us from obtaining a diverse list of
celebrity friends. For instance, almost all the top 100 images 
returned by `Beyonc\'e and' are with her husband Jay Z.
Therefore, to prevent such celebrity friends dominating 
the top returned results, a secondary search is conducted 
by explicitly removing pairs
found by the first-round search, \ie the query term now is
`\texttt{seed-celebrity} and \texttt{-friend1} \texttt{-friend2} ...',
where \texttt{friend1}, \texttt{friend2} ..\ are those found in the first search.
We do not perform more searches after the second round, as 
we find that the number of new pairs obtained in the third-round search 
is minimal.

\paragraph{Image download and filtering.}
Once the name pairs are obtained, we download the 
images and meta information from the image search engine, 
followed by name scanning again to remove false images. 
Face detection is performed afterwards to further refine 
the dataset by removing images with fewer than two faces.

\paragraph{De-duplication.} 
Removal of near duplicate images is performed 
similarly to~\citep{Parkhi15}, adapted to our case where images contain
multiple faces.
Namely, we extract a VLAD descriptor for 
each image (computed using densely extracted 
SIFT descriptors~\citep{Lowe04}) and
cluster all images for each celebrity.
This procedure results in per-celebrity clusters of near duplicate images.
Dataset-wide clustering is obtained as the union of all per-celebrity clusters,
where overlapping clusters
(\ie ones which share an image, which can happen because images contain
multiple celebrities) are merged.
Finally, de-duplication is performed by only keeping a single image
from each cluster.
Furthermore, images 
in common with the VGG Face Dataset~\citep{Parkhi15}
are removed as well.

\paragraph{Image annotation.}
Each face in each image is assigned to one of 2623 classes:
2622 celebrity names used to collect the dataset, or a special
``unknown person'' label if the person is not on the
predefined celebrity list; the ``unknown'' people are used as distractors.

As the list of celebrities is the same at that used in the VGG Face
Dataset~\citep{Parkhi15}, we use the pre-trained VGG-Face 
CNN~\citep{Parkhi15} to aid with annotation by using it to predict the identities of the faces in the images.  Combining the very confident
CNN classifications with our prior belief of who is depicted in the
image (\ie the query terms), results in many good quality automatic
annotations, which further decreases the manual annotation effort and
costs. But choosing thresholds for deciding which images need
annotation can be tricky, as we need to consider two cases: (i) if a
face in an image belongs to one of the query names
used for downloading that image, or (ii) if the face does not
belong to the query names.  The two cases require very different
thresholds, as explained in the following.
 
\paragraph{Identities in the query text.}
We first 
consider the case where the predicted face (using VGG-Face) matches the 
corresponding query words for downloading that image. 
In this case, there is a very strong prior that the predicted face is
correctly labelled as the predicted celebrity was explicitly searched for.
Therefore, if the face is scored higher than 0.2 by the CNN,
it is considered as being correctly labelled, otherwise it is sent for
human annotation.
As another way of including low scoring predictions,
if one of the query names is ranked within top 5 out of the 2622 
celebrities by the CNN, 
the face is also sent for annotation.
A face that does not pass any of the automatic or manual annotation
requirements, is considered to be an ``unknown'' person.
 
\paragraph{Identities not in the query text.}
The next question is: if a face is predicted to be one 
of the 2622 celebrities with a high confidence (CNN score) 
but it does not match any query celebrities, can we 
automatically label it as a celebrity without human 
annotation? In this situation, the CNN is much less 
likely to be correct as the predicted celebrity was not explicitly searched for.
Therefore, it has to pass a much stricter CNN score threshold of
0.8 in order to be considered to be correctly labelled without need for
manual annotation.
On the other hand, empirically we find that a prediction scored below 0.4
is always wrong, and is therefore automatically assigned the ``unknown'' label.
The remaining case, a face with score between 0.4 and 0.8, should be
manually annotated, but to save time and human effort we simply remove the
image from the dataset.

\begin{table}[t]
   \centering
\caption{\textbf{Distribution of faces per image 
in the `Celebrity Together' Dataset.}}
   \begin{tabular}{c||c|c|c|c|c}
      No. faces / image & 2   & 3  & 4  & 5 & \textgreater 5 \\ \hline
      No. images       & 113k & 43k & 19k & 9k & 10k             \\ %
   \end{tabular}
   \vspace{-0mm} 
   \label{tab:facehist}
\end{table}

\begin{table}[t]
   \centering
\caption{ \textbf{Distribution of annotations per image
   in the `Celebrity Together' Dataset.}}
   \begin{tabular}{c||c|c|c|c|c|c}
      No. celeb / image & 1  & 2  & 3  & 4  & 5 & \textgreater 5 \\ \hline
      No. images     & 88k & 89k & 12k & 3k & 0.7k & 0.3k           \\ %
   \end{tabular}
   \vspace{-0mm} 
   \label{tab:celehist}
\end{table}

\section{Experiments and results}\label{sec:exp}
In this section we investigate four aspects:
first, in Sec.~\ref{sec:stress} we study the performance of different models 
(SetNet and baselines)
as the number of faces  per image  in the dataset  is increased. 
Second, we compare the performance of SetNet and the best baseline model
on the real-world {\em Celebrity Together} dataset
in Sec.~\ref{sec:DPS}.
Third, the trade-off between time complexity and set retrieval quality
is investigated in Sec.~\ref{sec:ranking}.
Fourth, in Sec.~\ref{sec:gram}, 
we demonstrate that SetNet learns to increase the 
descriptor orthogonality, which is good for aggregation.

Note,  in all the experiments, there is no overlap between 
the query identities used for testing  and the identities used for training the 
network, as the VGG Face Dataset~\citep{Parkhi15} (used for testing, e.g.\ 
for forming the {\em Celebrity Together} dataset) and 
the VGGFace2 Dataset~\citep{Cao18} (used for training) share no common identities.

\paragraph{Evaluation protocol.}
We use Normalized Discounted Cumulative Gain
\citep{Jarvelin02} (nDCG) 
to evaluate set retrieval performance, as it can measure
how well images containing {\em all} the query identities and also {\em subsets} of the queries
are retrieved.
For this measurement, images have different relevance,
not just a binary positive/negative label;
the relevance of an image is equal to the number of query identities
it contains.
We report nDCG@10 and nDCG@30, where nDCG@N is the nDCG for the ranked list
cropped at the top $N$ retrievals.
nDCG is computed as the ratio 
between DCG and the ideal DCG computed on the
ground-truth ranking, where DCG is defined as:
\begin{equation}
	DCG@N = \sum_{i=1}^{N}(2^{rel(i)}-1)/ \log_2 (i+1)
\label{eq:nDCG}
\end{equation}
\noindent where $rel(i)$ denotes the relevance of the $i$th retrieved image.
nDCGs are written
as percentages, so the scores
range between 0 and 100.

\subsection{Stress test}\label{sec:stress}
In this test, we aim to investigate how different models 
perform with increasing number of faces per set (image)
 in the test dataset. The effects of varying the number of faces
per set used for training are also studied.
One randomly sampled face example per 
query identity is used to query the dataset. 
The experiment is 
repeated 10 times using different face examples, 
and the nDCG scores are averaged.

\paragraph{Test dataset synthesis.}
A base dataset with 64k face sets of 2 faces each
is synthesized, using only the face images of labelled 
identities in the {\em Celebrity Together} dataset.
A random sample of 100 sets of 2 identities are used as queries,
taking care that the two queried celebrities do appear
together in some dataset face sets.
To obtain four datasets of varying difficulty,
0, 1, 2 and 3 distractor faces per set
are sampled from the unlabelled face images
in {\em Celebrity Together} Dataset,
taking care to include only true distractor people, 
\ie people who are not in the list of labelled identities in
the {\em Celebrity Together} dataset.
Therefore, all four datasets contain the
same number of face sets (64k) but have a different 
number of faces per set, ranging from 2 to 5.
Importantly, by construction, the relevance of each set to each query is
the same across all four
datasets, which makes the performance
numbers comparable across them.

\paragraph{Methods.}
The \emph{SetNet} models are
trained as described in Sec.~\ref{sec:training},
where the suffix \emph{`-2'} or \emph{`-3'} denotes
whether 2- or 3-element sets are used during training.
For SetNet, the optional suffix \emph{`+W'} means that 
the set descriptors are whitened. 
For example, \emph{SetNet-2+W} is a model trained with 2-element sets
and whitening.
Whereas for baselines, 
\emph{`+W'} indicates whether
the face descriptors have been whitened 
and L2-normalized before aggregation.
We further investigate the effect of performing 
whitening after aggregation for the baselines,
indicated as \emph{`+W (W after agg.)'}.
In this test, both SetNet and the baseline use 
ResNet-50 with 128-D feature output as the feature 
extractor network.

Baselines follow the same naming convention,
where the architectural difference from the SetNet is that the
aggregator block 
is replaced with average-pooling followed by L2-normalization,
as shown in Fig.~\ref{fig:cnn}
(\ie they use the same feature extractor network, data augmentation, optional whitening, \etc). 
The baselines are trained in the same manner as SetNet.
The exceptions are \emph{Baseline} and \emph{Baseline+W}, which simply use 
the feature extractor network
with average-pooling, but no training.
It is important to note that training the modified ResNet-50 
and SENet-50 networks 
on the VGGFace2 dataset~\citep{Cao18} provides 
very strong baselines (referred to as \emph{Baseline}).
Although the objective of this paper is not 
face classification, to demonstrate the performance 
of this baseline network, we test it 
on the public IARPA Janus Benchmark A (IJB-A dataset)~\citep{Klare15}.
Specifically, we follow the same test procedure for 
the 1:N Identification task described 
in~\citep{Cao18}.
In terms of the true positive identification rate (TPIR),
the ResNet-50 128-D network 
achieves 0.975, 0.993 and 0.994 for the 
top 1, 5 and 10 ranking respectively, which is on par with 
the state-of-the-art networks in~\citep{Cao18}.

For reference, an upper bound performance
(\emph{Descriptor-per-element}, see Sec.~\ref{sec:DPE} for details)
is also reported, where no aggregation is performed and all descriptors
for all elements are stored.

\begin{figure*}[t]
\vspace{-0mm}
\begin{minipage}{.55\textwidth}
   \begin{center}
	   \captionsetup[subfloat]{margin=-3cm}
	   \subfloat[]{\label{main:a}}\includegraphics[width=0.85\textwidth]{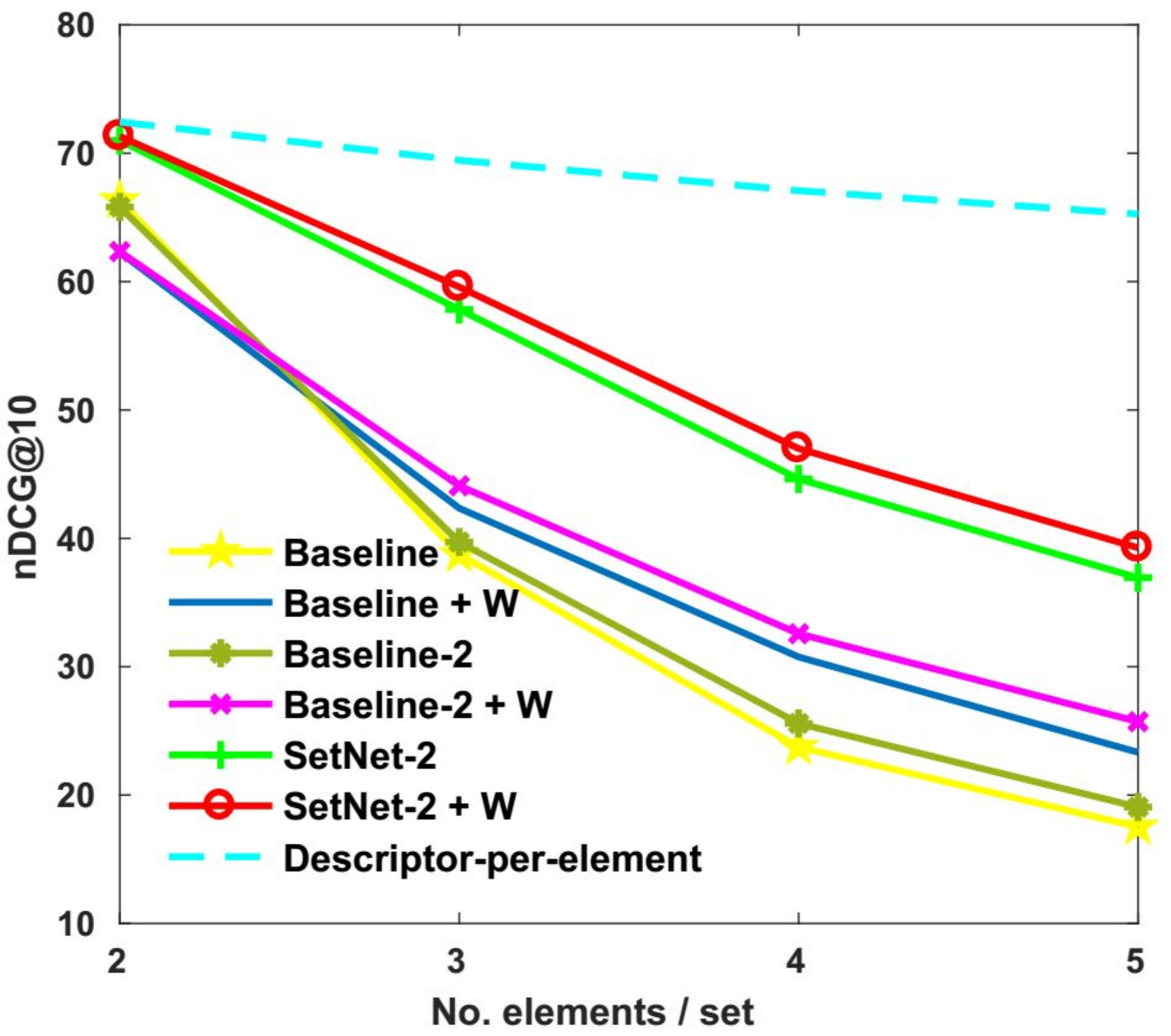}
   \end{center}
\end{minipage}\hfill
\begin{minipage}{.45\textwidth}
   \begin{center}
   \centering
	   \captionsetup[subfloat]{margin=-3cm}
	   \subfloat[][]{\label{main:b}
   \begin{tabular}{l@{~~~}|@{~~~}c@{~~~}c@{~~~}c@{~~~}c}
	   ~~Scoring model & 2/set         & 3/set      & 4/set  & 5/set       \\ \hline
	   Baseline    	& 66.3    & 38.8   & 23.7    & 17.5 \\
	   Baseline-2   & 65.8    & 39.7   & 25.6    & 19.0 \\ 
  	   Baseline-3    & 65.9    & 39.4   & 24.8    & 18.4 \\   
	   SetNet       & 71.1  & 55.1  & 43.2  & 34.9    \\
	   SetNet-2      & 71.0    & 57.7   & 44.6    & 36.9 \\  
	   SetNet-3      & 71.9    &  57.9  & 44.7    & 37.0 \\ 
           \hline
	   Baseline + W    & 62.3    & 42.3   & 30.8    & 23.3 \\ 
	   Baseline-2 + W   & 62.3    & 44.1   & 32.6    & 25.7 \\ 
	   Baseline-3 + W   & 62.1    & 44.0   & 32.3    & 25.6 \\ 
           Baseline-3 + W (W after agg.) & 62.2 & 44.0 & 32.2 & 25.6  \\
	   SetNet + W      & 71.2  & 57.5  & 45.7  & 37.8   \\
	   SetNet-2 + W     & 71.3    & 59.5   & 47.0    & 39.3 \\ 
	   SetNet-3 + W    & 71.8    & 59.8   & 47.1    & 39.3 \\ 
	   Desc-per-element  & 72.4    & 69.4   & 67.1    & 65.3 
   \end{tabular}
   }
   \end{center}
\end{minipage}\hfill
   \caption{ \textbf{Stress test  comparison of nDCG@10 of different models.} There are 100 query sets, each with two identities.
	   (a) nDCG@10 for different number of elements (faces) per set (image) in the test dataset.
	   (b) Table of nDCG@10 of stress test.
	   Columns corresponds to the four different test datasets
	defined by the number of elements (faces)  per set.
   }
   \label{fig:stress}
 \end{figure*}

\begin{figure*}[t]
\vspace{-0mm}
\begin{minipage}{.55\textwidth}
   \begin{center}
	   \captionsetup[subfloat]{margin=-3cm}
	   \subfloat[]{\label{main:a}}\includegraphics[width=0.85\textwidth]{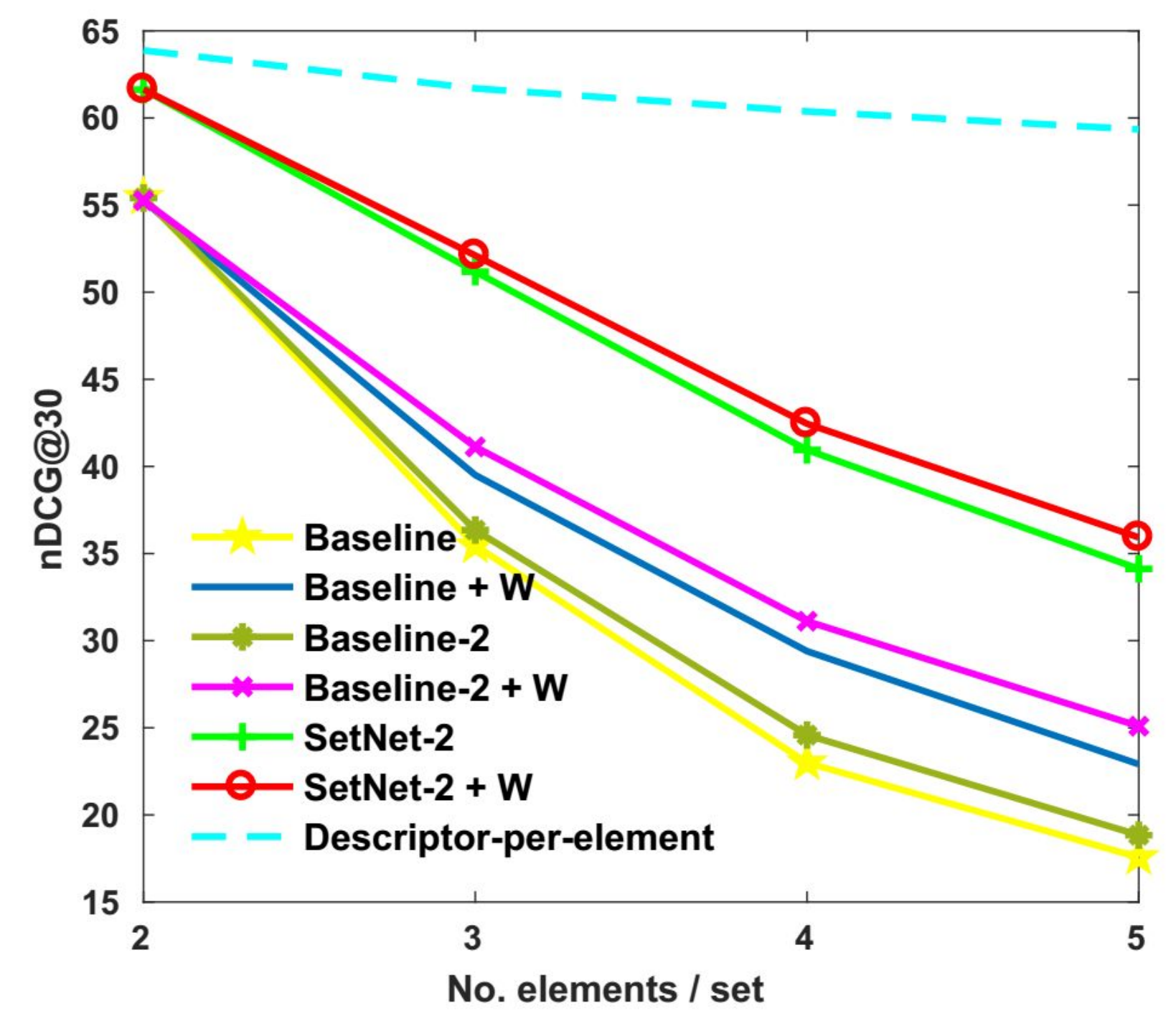}
   \end{center}
\end{minipage}\hfill
\begin{minipage}{.45\textwidth}
   \begin{center}
   \centering
	   \captionsetup[subfloat]{margin=-13cm}
	   \subfloat[][]{\label{main:b}
   \begin{tabular}{l@{~~~}|@{~~~}c@{~~~}c@{~~~}c@{~~~}c}
	   ~~Scoring model & 2/set         & 3/set      & 4/set  & 5/set       \\ \hline
	   Baseline    	& 55.5  & 35.5  & 23.0  & 17.6 \\
	   Baseline-2   & 55.4  & 36.3  & 24.6  & 18.8\\ 
  	   Baseline-3    & 55.3  & 35.9  & 23.9  & 18.3\\
           SetNet       & 61.1  & 49.2  & 39.1  & 32.2   \\
	   SetNet-2     & 61.6  & 51.1  & 40.9   & 34.1 \\
	   SetNet-3      & 61.5  & 51.2  & 41.0  & 34.2 \\
           \hline
	   Baseline + W    & 55.2  & 39.5  & 29.4  & 22.9\\
	   Baseline-2 + W   & 55.3  & 41.1  & 31.1  & 25.1 \\
	   Baseline-3 + W   & 55.2  & 41.0  & 30.9  & 25.0 \\
           Baseline-3 + W (W after agg.) & 55.2 & 41.0 & 30.9 & 25.0 \\
	   SetNet + W      & 61.4  & 50.5  & 40.7  & 34.5   \\
	   SetNet-2 + W     & 61.6  & 52.1  & 42.4  & 35.9\\
	   SetNet-3 + W    & 61.2  & 52.2  & 42.5  & 36.0 \\
	   Desc-per-element  & 63.9  & 61.7  & 60.4  & 59.3
   \end{tabular}
   }
   \end{center}
\end{minipage}\hfill
   \caption{ \textbf{Stress test comparison of nDCG@30 of different models.}
	   Columns corresponds to the four different test datasets
	defined by the number of elements (faces)  per set.
   }
   \label{fig:stress2}
 \end{figure*}

\paragraph{Results.}
From the results in Fig.~\ref{fig:stress} it is clear that \emph{SetNet-2+W} and \emph{SetNet-3+W}
outperform all baselines. 
As Fig.~\ref{fig:stress2} shows, nDCG@30 generally follows a
similar trend to nDCG@10: SetNet 
outperforms all the baselines by a large margin
and, moreover, the gap increases as the number of elements 
per set increases.
As expected, the performance 
of all models decreases as the number of elements per set 
increases due to larger cross-element interference.
However, \emph{SetNet+W} deteriorates more gracefully
(the margin between it and the baselines increases),
demonstrating that our training makes SetNet learn representations
which minimise the interference between elements.
The set training is beneficial for all architectures, 
including SetNet and baselines,
as models with suffixes \emph{`-2'} and \emph{`-3'}
(set training)
achieve better results than those without 
(no set training).

\noindent\textbf{Whitening} 
improves the performance for 
all architectures when there are more than 2 elements per set,  
regardless whether the whitening is applied before or after 
the aggregation.
This is a somewhat
surprising result since adding whitening only happens after the network
is trained. 
However, using whitening is common in the retrieval community
as it is usually found to be very helpful~\citep{Jegou12,Arandjelovic13},
but has also been used recently to improve CNN representations~\citep{Radenovic16,Sun17}.
\cite{Radenovic16} train a discriminative version of
whitening for retrieval, while
\cite{Sun17} reduce feature correlations for pedestrian retrieval
by formulating the SVD as a CNN layer.
It is likely that whitening before
aggregation is beneficial also because it makes descriptors more
orthogonal to each other, which helps to reduce the amount of
information lost by aggregation.  However, \emph{SetNet} gains much
less from whitening, which may indicate that it learns to produce
more orthogonal face descriptors.  
This claim is investigated further in Sec.~\ref{sec:gram}.

It is also important to note that,
as illustrated by Fig.~\ref{main:b},
the cardinality of the sets 
used for training does not affect the performance much,
regardless of the architecture.
Therefore,
training with a set size of 2 or 3 is sufficient
to learn good set representations which generalize
to larger sets.

\subsection{Evaluating on the Celebrity Together  dataset}\label{sec:DPS}
Here we evaluate the SetNet performance on the full
{\em Celebrity Together} dataset.

\paragraph{Test dataset.}
The dataset is described in Sec.~\ref{sec:dataset}.
To increase the retrieval difficulty, random 355k distractor images
are sampled from the MS-Celeb-1M Dataset~\citep{Guo16},
as before
taking care to include only true distractor people.
The sampled distractor sets are constructed such that the number of
faces per set follows the same distribution as in the {\em Celebrity Together} dataset.
The statistics of the resultant test dataset are shown in Table~\ref{tab:testset}.
There are 1000 test queries, formed by randomly sampling
500 queries containing two celebrities
and 500 queries containing three celebrities,
under the restriction that the queried celebrities do appear
together in some dataset images.

\paragraph{Experimental setup and baseline.}
In this test we consider two scenarios:
first, where only one face example is available 
for each query identity, and 
second, where three face examples 
per query identity are available. 
In the second scenario, for each query identity,
three face descriptors are extracted and aggregated
to form a single enhanced descriptor
which is then used to query the dataset.

In both scenarios the experiment 
is repeated 10 times using 
different face examples  for each query identity, and the nDCG score is averaged.
The best baseline from the stress test (Sec.~\ref{sec:stress})
\emph{Baseline-2+W} (modified ResNet with average-pooling),
is used as the main comparison method.
Moreover, we explore three additional variations of the baseline:
(i) average-pooling the L2 normalized features before (rather than after)
the final FC layer;
(ii) (\emph{`Baseline-GM'}), replacing the average-pooling with generalized mean pooling,
$g(p)= \left ( \frac{1}{n}\sum_{k=1}^{n} |x_k|^p \right )^{\frac{1}{p}}$, \citep{Rradenovic18} 
where the 
learnable parameter $p$ is  a
scalar shared across all feature dimensions;  and (iii)  (\emph{`Baseline-GM' (p per dim.)}),
replacing the average-pooling with generalized mean pooling
where the 
learnable parameter $p$ is  a vector with one value per dimension.
Following~\citep{Rradenovic18}, 
the learnable parameter $p$ is initialized to 3 for (ii)
and to a vector of 3's for (iii).
All three variations are trained 
with 2 elements per set.
The learnt $p$ for (ii) is about 1.43. For (iii), the elements in the learnt $p$ have 
various values mostly between $1.3$ and $1.5$.
Note that for (ii) and (iii), whitening happens after the aggregation due to
the restriction of the non-negative values in the descriptors.

\begin{table}[t]
\vspace{-0mm}
   \centering
\caption{
    \textbf{Number of images and faces in the test dataset.}
   The test dataset consists of the \emph{Celebrity Together} 
   dataset and distractor images from 
   the MS-Celeb-1M dataset~\citep{Guo16}.
   }
    \begin{tabular}{l||c|c|c}
           & Celebrity Together & \begin{tabular}[c]{@{}c@{}}Distractors \\ from MS1M\end{tabular} & Total \\ \hline
	    Images & 194k               & 355k                                       & 549k  \\ 
	    Faces  & 546k               & 1M   & 1546k
   \end{tabular}
   \label{tab:testset}
   \centering
\end{table}

\begin{table*}[t]
\centering
\caption{
	   \textbf{Set retrieval performance 
	on the test set.}
$D_e$ is the output feature dimension of the feature 
extractor CNN, and $D$ is the dimension of the output 
feature dimension of SetNet.
$Q$ is the number of identities in the query. There are 500 queries with $Q=2$,  and 500 
                   with  $Q=3$.
$N_{ex}$ is the number of 
	available face examples for each identity.
	$N_{qd}$ is the number of 
	descriptors actually used for querying.
	   }
   \begin{tabular}{l|c|c|c|c|cc|cc}
	   ~~~~~~~~Scoring model   & Feature & $D_{e}$ & $D$  & $N_{qd}$   &  \multicolumn{2}{c|}{Scenario 1: $N_{ex} = 1$}  &   \multicolumn{2}{c}{Scenario 2: $N_{ex} = 3$}     \\    
	                         & extractor &  &  &                   & {nDCG@10}       & {nDCG@30}   & {nDCG@10}       & {nDCG@30}  \\ \hline

	   Baseline-2 + W    & ResNet  & 128 & -       & $Q$    & 50.0          &   49.4    & 56.6          & 56.0     \\ 	   
	   Baseline-2 + W (agg. before FC)      & ResNet  & 128 & - & Q  & 50.2 & 49.5 & 56.6 & 56.0 \\ 
           Baseline-GM-2 + W     & ResNet  & 128 & - & Q  & 51.0 & 50.6 & 57.2 & 56.9 \\ 
           Baseline-GM-2 + W ($p$ per dim.)    & ResNet  & 128 & - & Q  & 50.9 & 50.4 & 57.0 & 56.8 \\ 

	   SetNet-3 + W      & ResNet   & 128 & 128         & $Q$    & 59.1          & 59.4     & \textbf{63.8}          & 64.1       \\ 
	   SetNet-3 + W w/ query agg.\  & ResNet  & 128 & 128   & 1   & 58.7    & 59.4    & 62.9    & 64.1 \\ 
	   \hline
	   Baseline-2 + W    & SENet  & 128 & -       & $Q$    & 52.9         & 52.5     &  59.6        & 59.5       \\ 	   
	   SetNet-3 + W      & SENet   & 128 & 128         & $Q$    & \textbf{59.5}         & \textbf{59.5}     & 63.7        &  \textbf{64.5}          \\ 
	   SetNet-3 + W w/ query agg.\  & SENet  & 128 & 128   & 1   &  59.1   &  \textbf{59.5}   & 63.4    & 64.0  \\ \hline

	   Baseline-2 + W    & ResNet  & 256 & -       & $Q$    & 56.8         & 57.8    & 62.5        & 64.2        \\ 	   
	   SetNet-3 + W      & ResNet   & 256 & 256         & $Q$    & 61.2         & 62.6   &  66.0        & 68.9            \\ 
	   SetNet-3 + W w/ query agg.\  & ResNet  & 256 & 256   & 1   & 60.3  & 62.4     & 65.1   & 68.8  \\ 
	   \hline
	   Baseline-2 + W    & SENet  & 256 & -       & $Q$    & 58.2         & 60.8   & 64.2   & 66.2      \\ 	   
	   SetNet-3 + W      & SENet   & 256 & 256         & $Q$    & \textbf{62.6}        & \textbf{64.8}   & \textbf{67.3}         & \textbf{70.1}        \\ 
	   SetNet-3 + W w/ query agg.\  & SENet  & 256 & 256   & 1   & 61.9   & 64.5    & 66.3    & 69.6   \\ \hline

   \end{tabular}
   \label{tab:results_DPS}
\end{table*}

\paragraph{Results.}
Table \ref{tab:results_DPS} shows that SetNets
outperform the best baseline for all performance measures
by a large margin, across different feature extractors
(\ie ResNet and SENet) and different feature dimensions 
(\ie 128 and 256).
The boost is particularly 
impressive when 
ResNet with 128-D output is used and
only one face example is available for each query identity,
where \emph{Baseline-2+W} is beaten by 9.1\% and 10.0\% at
nDCG@10 and nDCG@30 respectively.
As we can see, performing the average-pooling before the FC layer in 
the feature extractor brings a marginal increase in
the performance for the \emph{Baseline-2+W}.
A slightly larger enhancement is achieved by replacing the average-pooling in \emph{Baseline-2+W}
by the generalized mean with a learnable parameter $p$. 
In practice, we find that using a shared value of 
$p$ for all dimensions of the descriptors results in the best baseline model
\ie \emph{Baseline-GM-2+W}.
Although the baseline method is improved by a better (and learnable) 
mean computation method, the gap
between the baseline models and SetNet is still significant.

Similar improvement appears for SENet-based SetNet: 
6.6\% at nDCG@10 and 7\% at nDCG@30.
When a feature dimension of 256 is used, we also 
observe that \emph{SetNet-3+W} outperforms \emph{Baseline-2+W}
by a significant amount, \eg 4.4\% at nDCG@10 on both 
ResNet and SENet.
Note that the gap between the baseline and SetNet is smaller
when we use 256-D set-level descriptors rather than 128-D. 
This indicates that the superiority of SetNet is more obvious
with lower dimensionality.
The improvement is also significant for the second scenario where  three face examples 
are available for each query identity. 
For example, an 
improvement of 7.2\% and 8.1\% over the baseline is observed
on ResNet with 128-D output, and 4.1\% and 5\% on SENet.
We can also conclude that SENet achieves better results 
than ResNet as it produces better face descriptors.
In general, the results demonstrate that our trained aggregation method 
is indeed beneficial since it is designed 
and trained end-to-end exactly for the task in hand.
Fig.~\ref{fig:teaser} shows 
the top 3 retrieved images out of 549k images 
for two examples queries
using
SetNet (images are cropped for better viewing).

\paragraph{Query aggregation.}
We also investigate a more efficient method to 
query the database for multiple identities. 
Namely, we aggregate the 
descriptors of all the query identities using 
SetNet to produce a 
single descriptor which
represents all query identities, 
and query with this single descriptor.
In other words, we treat all the query images as 
a set, and then obtain a set-level query descriptor
using SetNet.
In the second scenario, when 
three face examples are available for each query identity, 
all of the descriptors are simply fed to SetNet to 
compute a single 
descriptor. With this query representation 
we obtain a slightly lower nDCG@10 
compared to the original method shown 
in Table~\ref{tab:results_DPS}
($62.9$ vs $63.8$), 
and the same nDCG@30 ($64.1$).
However, as will be seen in the next section,
this drop can be nullified by re-ranking,
making \emph{query aggregation} an 
attractive method due to its efficiency.
In the second scenario, apart from aggregating  
the query face descriptors, 
we also investigated other ways of making use of
multiple face examples per query, including 
scoring the dataset images with each face example 
separately followed by merging the scores for each
image under some combination rules (\eg mean, max, \etc).
However, it turns out that simply aggregating the 
face descriptors for each query identity achieves the best 
performance and, moreover, it is the most efficient 
method as it adds almost  no computational cost to the 
single-face-example scenario.

\subsection{Efficient set retrieval}
\label{sec:ranking}

Our SetNet approach stores a single descriptor-per-set making it very fast
though with potentially sacrificed accuracy.
This section introduces some alternatives and evaluates
trade-offs between set retrieval quality and retrieval speed.
To evaluate computational efficiency formally with the big-O notation,
let $Q$, $F$ and $N$ be the number of query identities, average number of faces
per dataset  image, and the number of dataset  images, respectively,
and let the face descriptor be $D$-dimensional.
Recall that our SetNet produces a compact set representation
which is also $D$-dimensional, and $D=128$ or $256$.

\paragraph{Descriptor-per-set (SetNet).}
Storing a single descriptor per set is very computationally efficient
as ranking only requires computing a scalar product between
$Q$ query $D$-dimensional descriptors
and each of the $N$ dataset descriptors,
passing them through a logistic function,
followed by scoring the images by the sum
of similarity scores, making this step $O(NQD)$.
For the more efficient \emph{query aggregation} where only one 
query descriptor is used to represent all the query
identities, this step is even faster with $O(ND)$.
Sorting the scores is $O(N \log N)$.
Total memory requirements are $O(ND)$.

\paragraph{Descriptor-per-element.}
\label{sec:DPE}
Set retrieval can also be performed by storing all element descriptors,
requiring $O(NFD)$ memory.
Assuming there is no need 
for handling strange (\eg Photoshopped) images, 
each query identity can be portrayed at most once in an image, 
and each face can only correspond to a single query identity. 
An image can be scored by obtaining all $Q \times F$ pairs of
(query-identity, image-face) scores and finding the optimal assignment
by considering it as a maximal weighted matching problem in a bipartite graph.
Instead of solving the problem using the Hungarian algorithm which has
computational complexity that is cubic in the number of faces and
is thus prohibitively slow, we use a greedy matching approach.
Namely, all (query-identity, image-face) matches 
are considered in decreasing order of similarity 
and added to the list of accepted matches if neither 
of the query person nor the image face have been 
added already. 
The complexity of this 
greedy approach is $O(QF\log(QF))$ per image.
Therefore, the total computational complexity is
$O(NQFD + NQF\log(QF) + N\log N)$.
For our problem, we do not find any loss in retrieval performance
compared to optimal matching, while being $7 \times$ faster.

\paragraph{Combinations by re-ranking.}
Borrowing ideas again from image retrieval~\citep{Sivic03,Philbin07},
it is possible to combine the speed benefits of the faster methods
with the accuracy of the slow descriptor-per-element method by
using the former for initial ranking, and the latter to re-rank
the top $N_r$ results. 
The computational complexity is then equal to that of
the fast method of choice, plus $O(N_r QFD + N_r QF\log(QF) + N_r\log N_r)$.

\paragraph{Pre-tagging.}
A fast but naive approach to set retrieval is to pre-tag all the dataset images 
offline with a closed-world
of known identities by deeming a person to appear in the image
if the  score is larger than a threshold. 
Set retrieval is then performed by ranking images 
based on the intersection between the query and image tags.
Pre-tagging seems straightforward, 
however, it suffers from two large limitations; 
first, it is very constrained as it only 
supports querying for a fixed set of identities 
which has to be predetermined at the offline tagging stage. 
Second, it relies on making hard decisions 
when assigning an identity to a face, 
which is expected to perform badly, 
especially in terms of recall.
Note to obtain high precision (\ie correct) 
results for tagging, 
10 face examples are used for making the decision
on whether each predetermined identity is in the 
database images.
Thus tagging is a somewhat unfair baseline as all the other methods
operate in a more realistic scenario of only having one to three examples
available.
The computational complexity is $O(NQ + N\log N)$ but pre-tagging is very
fast in practice as querying can be implemented efficiently using an inverted index.

\paragraph{Experimental setup.}
The performance is evaluated 
on the 1000 test queries
and on the same full
dataset with distractors as in Sec.~\ref{sec:DPS}.
$N_r$ is varied in this experiment to demonstrate the trade-off between accuracy and 
retrieval speed.
Here, there is one available face example for each identity, $N_{ex}=1$.
For the descriptor-per-element and pre-tagging methods, 
we use the Baseline + W features.

\begin{table*}[t!]
\vspace{-0mm}
   \centering
  \caption{
       \textbf{Retrieval speed \emph{vs} quality trade-off 
	   with varied number of re-ranking images.}
	   Retrieval performance, average time required to execute a set query
	   and speedup over \emph{Desc-per-element} are shown
	   for each method. `Re.' denotes re-ranking,  
	   $D$ denotes the dimension of both 
	   the set-level descriptors and the descriptors used 
	   in pre-tagging method,
	   and $N_r$ denotes the number of re-ranked images.
           The evaluation is on the 1000 test queries and on the same full dataset with distractors as in Sec.~\ref{sec:DPS}.
           $^\dagger$ Note that \emph{Pre-tagging} is an unfair baseline, as explained in Sec.~\ref{sec:ranking}.
   }
   \begin{tabular}{l@{~~}c|c@{~}c|c|c@{~~}c@{~~}c@{~~}c|c@{~~}c@{~~}c@{~~}c}

	   ~~Scoring method & Architecture & Feature & $D$ & $N_r$  & \multicolumn{4}{c}{Without query aggregation} &\multicolumn{4}{c}{With query aggregation}\\
           & & extractor &  & & nDCG & nDCG & & & nDCG & nDCG & & \\
	   &  & & &       & {@10}       & {@30}       & {Timing} & {Speedup}  & {@10}       & {@30}       & {Timing} &{Speedup}  \\ 
	   \hline
	   Desc-per-set  & Baseline-GM-2 + W                   & ResNet & 128 & -         & 51.0 & 50.6   & 0.11s & $57.5\times$ & 50.6 & 50.6 & 0.01s & $635\times$ \\        
	   Desc-per-set + Re. & Baseline-GM-2 + W               & ResNet & 128 & 2000         & 78.5           & 76.2  & 0.28s & $22.7\times$ & 78.5 & 76.2 & 0.03s & $212\times$ \\    
	   Desc-per-element & Pre-tagging$^\dagger$ 	 & ResNet & 128	& -         &   65.7         & 68.6   & 0.01s & $635\times$ & - & - & - & - \\        
	   Desc-per-element + Re. & Pre-tagging$^\dagger$ 	 & ResNet & 128	& 2000         &  84.6          & 82.4   & 0.18s & $35.3\times$ & - & - & - & - \\        \hline

	   Desc-per-set 	& SetNet-3 + W & ResNet & 128	& -         & 59.1           & 59.4   & 0.11s & $57.5\times$ & 58.7 & 59.4 & 0.01s & $635\times$ \\        
	   Desc-per-set + Re.   & SetNet-3 + W & ResNet & 128	& 100         & 76.5       & 70.1   & 0.13s & $48.8\times$ & 76.4 & 70.1 & 0.03s & $212\times$ \\     
	   Desc-per-set + Re.   & SetNet-3 + W & ResNet & 128	& 1000    &   84.2         & 80.0 & 0.20s & $31.8\times$ & 84.1 & 80.0 & 0.10s & $63.5\times$ \\        
	   Desc-per-set + Re.   & SetNet-3 + W & ResNet & 128	& 2000       &  85.3          & 81.4 & 0.28s & $22.7\times$ & 85.3 & 81.4 & 0.18s & $35.3\times$ \\       
	   Desc-per-element     & Baseline + W & ResNet & 128	& -      & 85.4           & 81.7     & 6.35s  & - & - & - & - & - \\ \hline

	   Desc-per-set + Re.   & SetNet-3 + W & SENet & 128 	& 2000  &  85.8          & 81.0        & 0.28s & $22.7\times$ & 85.8 & 81.0 & 0.18s & $35.3\times$ \\      
	   Desc-per-element     & Baseline + W & SENet & 128 	&  -     & 85.8           & 81.8     &  6.35s & - & - & - & - & - \\ 
	   Desc-per-set + Re. 	& SetNet-3 + W & ResNet & 256 &  2000        & 85.2           & 82.1   & 0.40s & $21.3\times$ & 85.2 & 82.1 & 0.19s & $44.9\times$ \\        
	   Desc-per-element     & Baseline + W & ResNet & 256 & -         & 85.6           & 82.3     & 8.53s  & - & - & - & - & - \\  
	   Desc-per-set + Re.   & SetNet-3 + W & SENet & 256 	& 2000        & 85.7           & 83.4  & 0.40s & $21.3\times$ & 85.7 & 83.4 & 0.19s & $44.9\times$ \\       
	   Desc-per-element     & Baseline + W & SENet & 256 	& -       & 85.9           & 83.5     & 8.53s  & - & - & - & - & - \\ 
   \end{tabular}
   \label{tab:hybrid}
\end{table*}

\paragraph{Speed test implementation.}
The retrieval speed is measured 
as the mean over all 1000 test query sets.
The test is implemented in Matlab, and the measurements 
are carried out on a Xeon E5-2667 v2/3.30GHz, 
with only a single thread.

\subsubsection{Results}
Table \ref{tab:hybrid} shows set retrieval results for the various methods
together with the time it takes to execute a set query.
The full descriptor-per-element approach is the most accurate one,
but also prohibitively slow for most uses, 
taking more than 6 seconds 
to execute a query using 128-D descriptors
(and 8 seconds for 256-D.)
The descriptor-per-set (\ie SetNet) 
approach with query aggregation is blazingly 
fast with only 0.01s per query
using one descriptor to represent all query identities,
but sacrifices retrieval quality to achieve this speed.
However, taking the 128-D descriptor as an example, 
using SetNet for initial ranking followed by re-ranking
achieves good results without a significant speed hit --
 the accuracy almost reaches that of 
the full slow descriptor-per-element 
(\eg the gap is 0.3\% for ResNet and 0.8\% for SENet at nDCG@30),
while being more than $35 \times$ faster.
A even larger speed gain is observed using 256-D descriptor,
namely about $45\times$ faster than the desc-per-element method.
Furthermore, by combining \emph{desc-per-set} and 
\emph{desc-per-element} it is possible to choose 
the trade-off between speed and retrieval quality, 
as appropriate for specific use cases. 
For a task where speed is crucial, \emph{desc-per-set} 
can be used with few re-ranking images (\eg 100) to obtain 
a $212\times$ speedup over the most accurate method
(\emph{desc-per-element}). 
For an accuracy-critical task, it is possible to 
re-rank more images while maintaining a 
reasonable speed. 

\paragraph{Baselines.}
The best baseline method \emph{Baseline-GM-2+W} 
is inferior to our proposed SetNet even when
combined with re-ranking. This is mainly because the 
initial ranking using \emph{Baseline-GM-2+W} does not have a large enough recall --
it does not retrieve a
sufficient number of target images in the top 2000 ranks,
so re-ranking is not able to reach the SetNet performance.

\paragraph{Pre-tagging.}
Pre-tagging achieves better results than 
\emph{desc-per-set} using SetNet descriptors without 
re-ranking.
This is reasonable as pre-tagging should be 
categorized into the desc-per-element method, \ie
it performs offline face recognition on individual face 
descriptors.
However, it is significantly lower than
SetNet even with only 100 re-ranking images,
\eg 65.7\% vs. 76.5\% in nDCG@10.
Lastly, for a more fair comparison, we perform 
re-ranking for the pre-tagging baseline. 
With the same number of re-ranked images (\ie 2000),
desc-per-set + re-ranking beats pre-tagging in nDCG@10
(85.3\% vs. 84.6\%) while having slightly lower nDCG@30.
Note that the pre-tagging method makes use of 10 face examples 
per identity for face recognition, whereas only one 
face example is available for each query identity for SetNet.
Most importantly, the pre-tagging method suffers 
from the limitation of retrieving only a closed-world set  of identities.
In contrast,  desc-per-set + re-ranking, is open-world as the query faces can simply 
be supplied at run-time. Thus it  is strongly preferred 
over pre-tagging in 
practice.

\begin{figure*}[t]
  \begin{center}
        \includegraphics[width=1\linewidth]{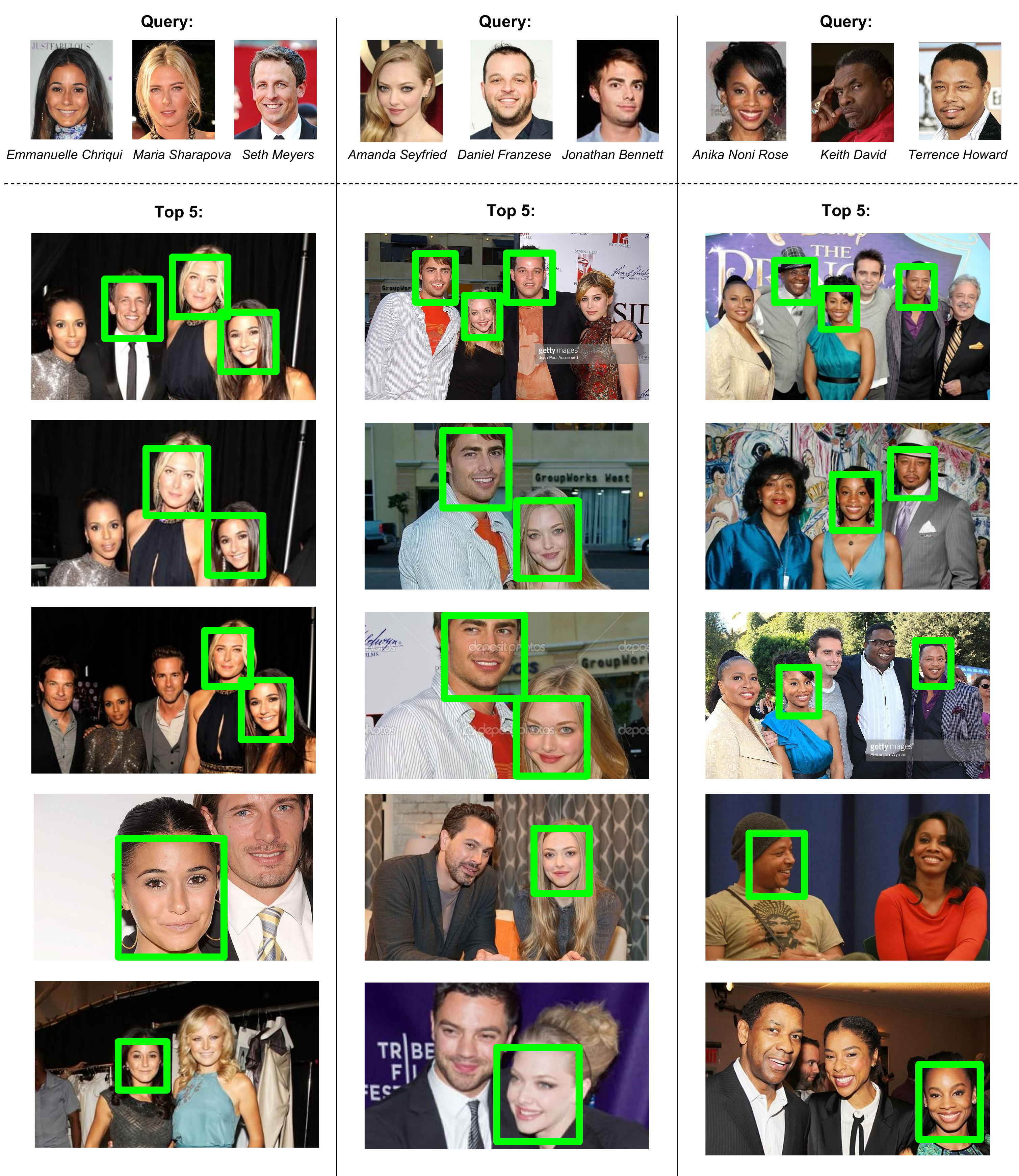}
  \end{center}
  \caption{\textbf{Top 5 ranked images returned 
	  by the SetNet based descriptor-per-set retrieval.}
 The query identities are highlighted by 
  green bounding boxes.}
\label{fig:rank1}
\end{figure*}

\begin{figure*}[t]
  \begin{center}
        \includegraphics[width=0.95\linewidth]{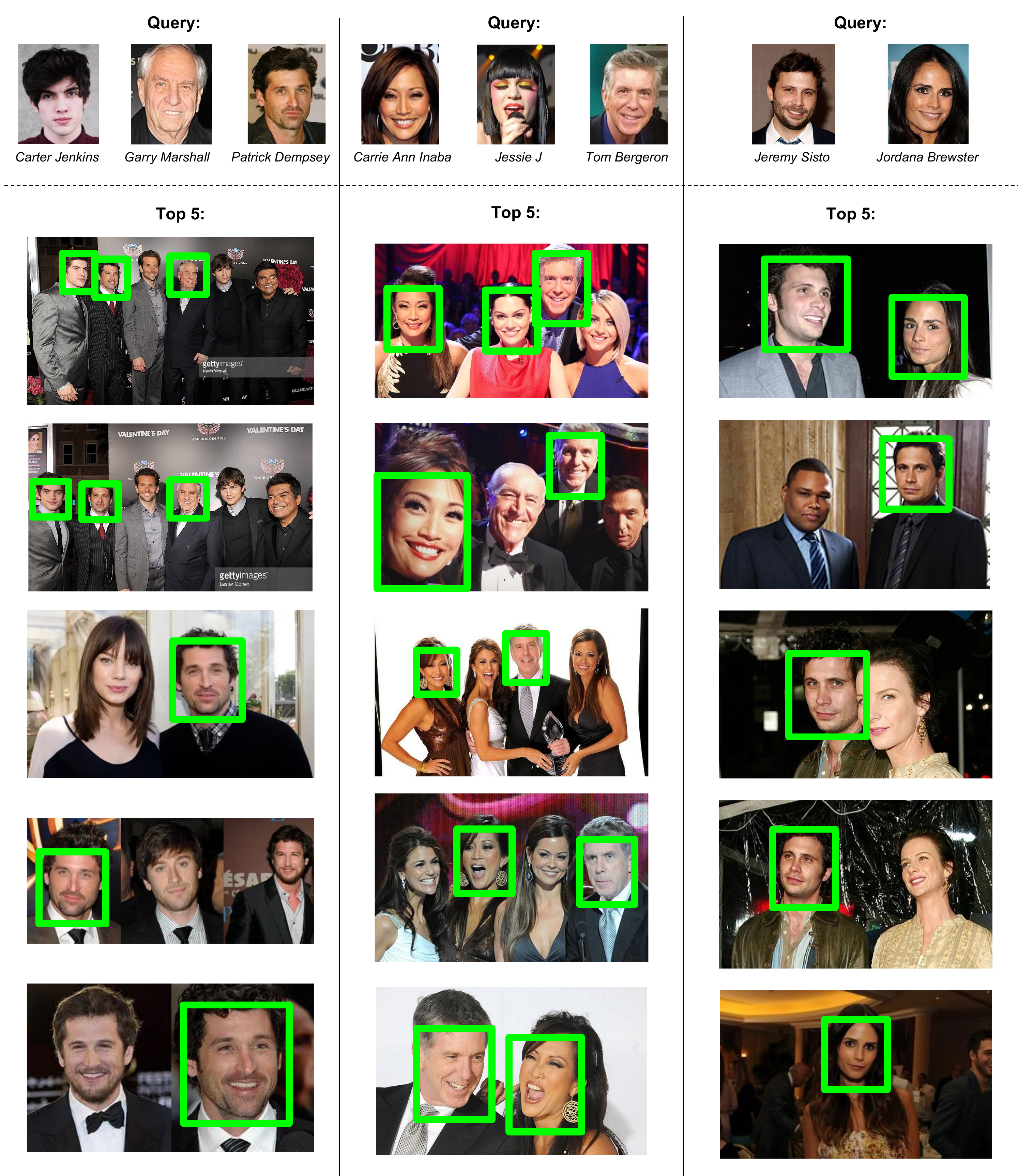}
  \end{center}
  \caption{\textbf{Top 5 ranked images returned 
	  by the SetNet based descriptor-per-set retrieval.}
 The query identities are highlighted by 
  green bounding boxes.}
\label{fig:rank2}
\end{figure*}

\begin{figure*}[t]
  \begin{center}
        \includegraphics[width=0.95\linewidth]{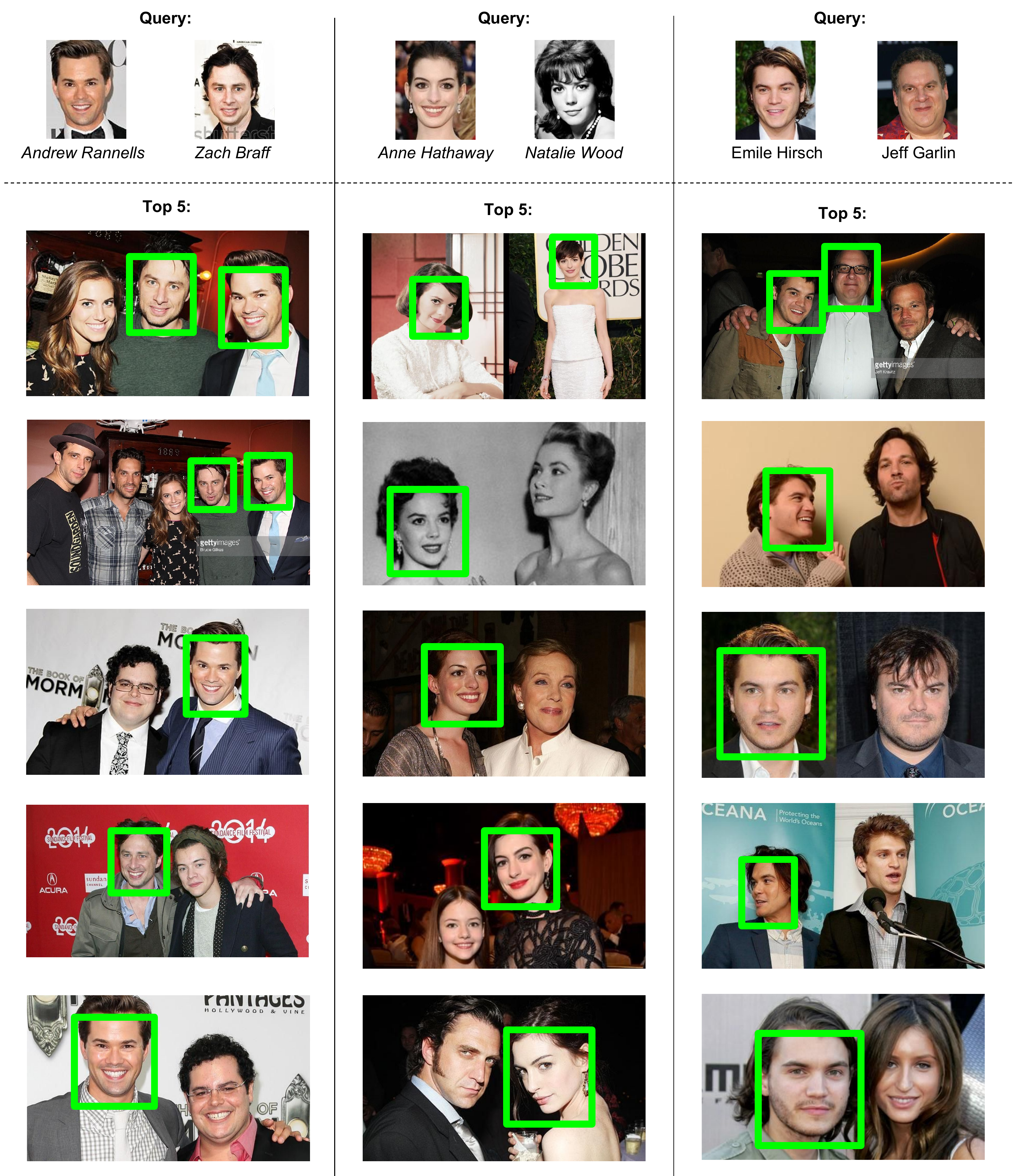}
  \end{center}
  \caption{\textbf{Top 5 ranked images returned 
	  by the SetNet based descriptor-per-set retrieval.}
 The query identities are highlighted by 
  green bounding boxes.}
\label{fig:rank3}
\end{figure*}

\paragraph{Retrieval examples.}
We visualize the performance of 
the set retrieval system based on descriptors 
produced by SetNet by showing some retrieval examples.
The top ranked 5 images (ordered column-wise) 
for several queries are shown in Fig.~\ref{fig:rank1}, \ref{fig:rank2},
\ref{fig:rank3}
(images are cropped for displaying purposes in order to see the faces better).
Notably, the nDCG@5 (the quality of the top 5 retrievals) is equal to 1 for all examples,
meaning that images are correctly ranked according to the size of their overlap with
the query set.
The retrieval is performed by desc-per-set + re-ranking 
using descriptors produced by a 128-D ResNet-based SetNet.

 \subsection{Descriptor orthogonality}\label{sec:gram}
As observed in Sec.~\ref{sec:stress}, whitening face descriptors before
aggregation improves the performance of baselines.
This section investigates this phenomenon further as well as the behaviour
of face descriptors learnt by SetNet.

We hypothesize that whitening helps because
it decorrelates
each dimension of the descriptors, and consequently descriptors will interfere less when
added to each other.
Furthermore,
we expect SetNet to automatically learn
such behaviour in order to minimize the 
interference between descriptors after aggregation. 
To verify this, we measure the extent to which face descriptors are
orthogonal to each other as follows.
We compute the Gram matrix of face descriptors for
2622 identities in the VGG Face Dataset
and measure the 
deviation $G_{diff}$ between the computed 
Gram matrix and an identity matrix (the Gram matrix of fully orthogonal descriptors):

\begin{equation}  
	G_{diff} = \left \| G - I \right \|_{Frob}
\label{eq:G}  
\end{equation}  
\noindent where $G$ is the Gram matrix and $I$ is the identity matrix.
Smaller $G_{diff}$ corresponds to more 
orthogonal descriptors.

\paragraph{Implementation details.}
$G_{diff}$ is computed in four steps: 
first, 100 faces are randomly sampled for each 
identity in the  VGG Face Dataset, 
and a feature vector for each face sample
is extracted by the network;
second, a single identity descriptor is obtained for each identity
by averaging the 100 face descriptors 
followed by L2-normalization;
third, 
the $2622 \times 2622$ Gram matrix is computed by 
taking the scalar product
between all possible pairs of identity descriptors
after mean subtraction and re-normalization.
We compute  $G_{diff}$ using non-whitened 
and whitened features of the 
best baseline network, \emph{Baseline-2}, and SetNet. 
Both networks are trained on the
VGGFace2 dataset~\citep{Cao18}, and their corresponding 
whitening transformations are computed on the same dataset.
To evaluate descriptor orthogonality,
the Gram matrix and $G_{diff}$ are computed using the independent
set of identities of VGGFace~\citep{Parkhi15}.

\paragraph{Results.}
As Table~\ref{tab:gramm} shows,
face features produced by \emph{SetNet-3} are much more orthogonal than
the ones from \emph{Baseline-2}, confirming our intuition that
this behaviour should naturally emerge from our training procedure.
Whitening reduces 
$G_{diff}$ for both \emph{Baseline-2} and \emph{SetNet-3},
as expected,
but the reduction is small for \emph{SetNet-3} as its descriptors are
already quite orthogonal, while it is large for \emph{Baseline-2} since
those are not.
Furthermore,
SetNet without whitening has a smaller $G_{diff}$ 
than even the whitened baseline, demonstrating that SetNet learns to produce
descriptors which are well suited for aggregation.

\begin{table}[t]
\center
	\caption{ 
	\textbf{Difference between Gram 
   	matrix and identity matrix $G_{diff}$.} The Gram matrix 
   	is computed over 2622 identities in VGG Face Dataset. 
        $D$ denotes the dimension of the 
 	set-level descriptors.
        }
\label{tab:gramm}
	\begin{tabular}{@{~~~}l@{~~~}|@{~~~}c@{~~~}c@{~~~}|@{~~~}c@{~~~}c@{~~~}}
		Model       & Feature extractor & $D$ & Not whitened    & Whitened    \\ \hline
		Baseline-2    	&  ResNet-50 & 128 & 399  &  276  \\
		SetNet-3   & ResNet-50 & 128 & 274  &  268
  	\end{tabular}
\end{table}

\section{Conclusion}
We have considered a new problem of set retrieval, discussed
multiple different approaches to tackle it, and evaluated
them on a specific case of searching for sets of faces.
Our learnt compact set representation,
produced using the SetNet architecture and trained in a novel manner directly
for the set retrieval task, beats all baselines convincingly.
Furthermore, due to its high speed it can be used for fast set retrieval
in large image datasets.
The set retrieval problem has applications beyond multiple faces in an image.
For example, a similar situation would also apply for the task of video retrieval when the elements
themselves are images (frames), and the set  is a video clip or shot.

We have also introduced a new dataset, {\em Celebrity Together},
which can be used to evaluate set retrieval performance  and to facilitate research on this new topic.
The dataset is available at \url{http://www.robots.ox.ac.uk/~vgg/data/celebrity_together/}.

\begin{acknowledgements}
 This work was funded by an EPSRC studentship and 
 EPSRC Programme Grant Seebibyte EP/M013774/1.
\end{acknowledgements}

\clearpage

\bibliographystyle{spbasic}      %
\bibliography{shortstrings,mybib,vgg_local,vgg_other}

\end{document}